\pdfoutput=1

\documentclass[11pt]{article}

\usepackage{acl}

\usepackage{times}
\usepackage{latexsym}

\usepackage[T1]{fontenc}

\usepackage[utf8]{inputenc}

\usepackage{microtype}

\usepackage{graphicx}

\title{On the Use of BERT for Automated Essay Scoring:\\ Joint Learning of Multi-Scale Essay Representation}

\author{
Yongjie~Wang$^1$~~~~Chuan~Wang$^1$~~~~Ruobing~Li$^1$~~~~Hui~Lin$^{1,2}$\\\\
$^1$LAIX Inc.\\
$^2$Shanghai Key Laboratory of Artificial Intelligence in Learning and Cognitive Science \\\\
\texttt{\{yongjie.wang, chuan.wang, ruobing.li, hui.lin\}@liulishuo.com} 
}

\date{}

\begin{document}
\maketitle
\begin{abstract}
In recent years, pre-trained models have become dominant in most natural language processing (NLP) tasks.
However, in the area of Automated Essay Scoring (AES), pre-trained models such as BERT have not been properly used to outperform other deep learning models such as LSTM.
In this paper, we introduce a novel multi-scale essay representation for BERT that can be jointly learned.
We also employ multiple losses and transfer learning from out-of-domain essays to further improve the performance.
Experiment results show that our approach derives much benefit from joint learning of multi-scale essay representation and obtains almost the state-of-the-art result among all deep learning models in the ASAP\footnote{\url{https://www.kaggle.com/c/asap-aes}} task.
Our multi-scale essay representation also generalizes well to CommonLit Readability Prize~(CRP\footnote{\url{https://www.kaggle.com/c/commonlitreadabilityprize/data}}) data set, which suggests that the novel text representation proposed in this paper may be a new and effective choice for long-text tasks.
\end{abstract}

\section{Introduction}

AES is a valuable task, which can promote the development of automated assessment and help teachers reduce the heavy burden of assessment.
With the rise of online education in recent years, more and more researchers begin to pay attention to this field.

AES systems typically consist of two modules, which are essay representation and essay scoring modules. The essay representation module extracts features to represent an essay and the essay scoring module rates the essay with the extracted features.

When a teacher rates an essay, the scores are often affected by multiple signals from different granularity levels, such as token level, sentence level, paragraph level and etc.
For example, the features may include the numbers of words, the essay structure, the master degree of vocabulary and syntactic complexity, etc.
These features come from different scales of the essay.
This inspires us to extract multi-scale features from the essays which represent multi-level characteristics of the essays.

Most of the deep neural networks AES systems use LSTM or CNN. Some researchers~\citep{Uto:2020, Rodriguez:2019, Mayfield:2020} attempt to use BERT~\citep{Devlin:2019} in their AES systems but fail to outperform other deep neural networks methods~\citep{Dong:2017, Tay:2018}.
We believe previous approaches using BERT for AES suffer from at least three limitations. First, the pre-trained models are usually trained on sentence-level, but fail to learn enough knowledge of essays. Second, the AES training data is usually quite limited for direct fine-tuning of the pre-trained models in order to learn better representation of essays.
Last but not least, mean squared error (MSE) is commonly used in the AES task as the loss function.
However, the distribution of the sample population and the sorting properties between samples are also important issues to be considered when designing the loss functions as they imitate the psychological process of teachers rating essays.
Different optimizations can also bring diversity to the final overall score distribution and contribute to the effectiveness of ensemble learning.

To address the aforementioned issues and limitations, we introduce joint learning of multi-scale essay representation into the AES task with BERT, which outperforms the state-of-the-art deep learning models based on LSTM ~\citep{Dong:2017, Tay:2018}.
We propose to explicitly model more effective representations by extracting multi-scale features as well as leveraging the knowledge learned from numerous sentence data.
As the training data is limited, we also employ transfer learning from out-of-domain essays which is inspired by~\citep{Song:2020}.
To introduce the diversity of essay scoring distribution, we combine two other loss functions with MSE.
When training our model with multiple losses and transfer learning using R-Drop~\citep{Liang:2021}, we almost achieve the state-of-the-art result among all deep learning models. The source code of prediction module with a trained
model for ASAP's prompt 8 is publicly available\footnote{\url{https://github.com/lingochamp/Multi-Scale-BERT-AES}}.

In summary, the contribution of this work is as follows: 
\begin{itemize}
\item We propose a novel essay scoring approach to jointly learn multi-scale essay representation with BERT, which significantly improve the result compared to traditionally using pre-trained language models. 
\item Our method shows significant advantages in long text tasks and obtains almost the state-of-the-art result among all deep learning models in the ASAP task.
\item We introduce two new loss functions which are inspired by the mental process of teacher rating essays, and employ transfer learning from out-of-domain essays with R-Drop~\citep{Liang:2021}, which further improves the performance for rating essays. 
\end{itemize}

\section{Related Work}

The dominant approaches in AES can be grouped into three categories: traditional AES, deep neural networks AES and pre-training AES.

\begin{itemize}
\item \textbf{Traditional AES} usually uses regression or ranking systems with complicated handcrafted features to rate an essay~\citep{Larkey:1998,  Rudner:2002, Attali:2006, Yannakoudakis:2011, Chen:2013, Phandi:2015, Cozma:2018}.
These handcrafted features are based on the prior knowledge of linguists.
Therefore they can achieve good performance even with small amounts of data.

\item \textbf{Deep Neural Networks AES} has made great progress and achieved comparable results with traditional AES recently ~\citep{Taghipour:2016, Dong:2016, Dong:2017, Alikaniotis:2016, Wang:2018, Tay:2018, Farag:2018, Song:2020, Ridley:2021,  Muangkammuen:2020, Mathias:2020}.
While the handcrafted features are complicated to implement and careful manual design makes these features less portable, deep neural networks such as LSTM or CNN can automatically discover and learn complex features of essays, which makes AES an end-to-end task.
Saving much time to design features, deep neural networks can transfer well among different AES tasks.
By combining traditional and deep neural network approaches, AES can even obtain a better result, which benefits from both representations~\citep{Jin:2018, Dasgupta:2018, Uto:2020}.
However, ensemble way still needs handcrafted features which cost numerous energy of researchers.

\item \textbf{Pre-training AES} uses the pre-trained language model as the initial essay representation module and fine-tune the model on the essay training set.
Though the pre-trained methods have achieved the state-of-the-art performance in most NLP tasks, most of them ~\citep{Uto:2020, Rodriguez:2019, Mayfield:2020} fail to show an advantage over other deep learning methods ~\citep{Dong:2017, Tay:2018} in AES task. As far as we know, the work from ~\citet{Cao:2020} and ~\citet{Yang:2020} are the only two pre-training approaches which surpass the other deep learning methods. Their improvement mainly comes from the training optimization. ~\citet{Cao:2020} employ two self-supervised tasks and domain adversarial training, while ~\citet{Yang:2020} combine regression and ranking to train their model.

\end{itemize}


\section{Approach}

\subsection{Task Formulation}
The AES task is defined as following:

Given an essay with $n$ words $X=\{x_i\}^n_{i=1}$, we need to output one score $y$ as a result of measuring the level of this essay.

Quadratic weighted Kappa (QWK)~\citep{Cohen1968} metric is commonly used to evaluate AES systems by researchers, which measures the agreement between the scoring results of two raters.

\subsection{Multi-scale Essay Representation}

We obtain the multi-scale essay representation from three scales: token-scale, segment-scale and document-scale.

\textbf{Token-scale and Document-scale Input}
We apply one pre-trained BERT~\citep{Devlin:2019} model for token-scale and document-scale essay representations. The BERT tokenizer is used to split the essay into a token sequence $T_{1}=[t_1, t_2, ...... t_n]$, where $t_i$ is the $i$th token and $n$ is the number of the tokens in the essay.
The \textbf{token} we mentioned in this paper all refer to WordPiece, which is obtained by the subword tokenization algorithm used for BERT.
We construct a new sequence $T_{2}$ from $T_{1}$ as following. $L$ is set to 510, which is the max sequence length supported by BERT except the token $[CLS]$ and $[SEP]$.

\begin{small}
$T_2=
\left\{
\begin{array}{lcl}
$[CLS]$ + [t_1, t_2, .., t_{L}] + $[SEP]$    &&{n > L}\\
$[CLS]$ + T_1 + $[SEP]$ &&{n=L}\\
$[CLS]$ + T_1 + $[PAD]$ * (L-n) + $[SEP]$ &&{n<L}\\
\end{array} \right . $
\end{small}

The final input representation are the sum of the token embeddings, the segmentation embeddings and the position embeddings.
A detailed description can be found in the work of BERT \citep {Devlin:2019}.

\textbf{Document-scale}
The document-scale representation is obtained by the $[CLS]$ output of the BERT model. As the $[CLS]$ output aggregates the whole sequence representation, it attempts to extract the essay information from the most global granularity.
   
\textbf{Token-scale}
As the BERT model is pre-trained by Masked Language Modeling~\citep{Devlin:2019}, the sequence outputs can capture the context information to represent each token.
An essay often consists of hundreds of tokens, thus RNN is not the proper choice to combine all the token information due to the gradients vanishing problem.
Instead, we utilize a max-pooling operation to all the sequence outputs and obtain the combined token-scale essay representation.
Specifically, the max-pooling layer generates a $d$-dimensional vector ${W=[w_1, w_2, ..., w_j, ..., w_{d}]}$ and the element $w_j$ is computed as below:
\begin{center}
$w_j = max\{h_{1, j}, h_{2, j}, ..., h_{n, j}\}$
\end{center}
where $d$ is the hidden size of the BERT model.
As we use the pre-trained BERT model \textbf{bert-base-uncased}\footnote{\url{https://huggingface.co/bert-base-uncased}}, the hidden size $d$ is 768.
All the $n$ sequence outputs of the BERT model are annotated as $[h_1, h_2, ..., h_i, ...,  h_n]$, where $h_i$ is a $d$-dimensional vector $[h_{i,1}, h_{i,2}, ..., h_{i,d}]$ representing the $i$th sequence output, and $h_{i, j}$ is the $j$th element in $h_i$.

\textbf{Segment-scale} Assuming the segment-scale value set is $K=[k_1, k_2, ... k_i, ..., k_S]$, where $S$ is the number of segment scales we want to explore, and $k_i$ is the $i${th} segment-scale in $K$.
Given a token sequence $T_{1}=[t_1, t_2, ...... t_n]$ for an essay, we obtain the segment-scale essay representation corresponding to scale $k_i$ as follows:
\begin{enumerate}
\item 
We define $n_p$ as the maximum number of tokens corresponding to each essay prompt $p$.
We truncate the token sequence to $n_p$ tokens if the essay length is longer than $n_p$, otherwise we pad $[PAD]$ to the sequence to reach the length $n_p$.


\item Divide the token sequence into $m = \lceil{n_p/k_i}\rceil$ segments and each segment is of length $k_i$ except for the last segment, which is similar to the work of ~\citep{Mulyar:2019}.
 
\item Input each of the $m$ segment tokens into the BERT model, and get $m$ segment representation vectors from the $[CLS]$ output.

\item Use an LSTM model to process the sequence of $m$ segment representations, followed by attention pooling operation on the hidden states of the LSTM output to obtain the segment-scale essay representation corresponding to scale $k_i$.
\end{enumerate}
  
The LSTM cell units process the sequence of segment representations and generate the hidden states as follows:

\begin{center}
\label{eq:lstm_it}
$i_t =\sigma(Q_i \cdot s_t + U_i \cdot {h}_{t-1} + b_i)$ \\
\label{eq:lstm_ft}
$f_t =\sigma(Q_f \cdot s_t + U_f \cdot {h}_{t-1} + b_f)$  \\
\label{eq:lstm_ct_hat}
$\hat c_t =tanh(Q_c \cdot s_t + U_c \cdot {h}_{t-1} + b_c)$  \\
\label{eq:lstm_ct}
$c_t =i_t \circ \hat c_t + f_t \circ {c}_{t-1}$  \\
\label{eq:lstm_ot}
$o_t =\sigma(Q_o \cdot s_t + U_o\cdot {h}_{t-1} + b_o)$ \\
\label{eq:lstm_ht}
$h_t = o_t \circ tanh(c_t)$  \\
\end{center}

 where $s_t$ is the $t${th} segment representation from BERT $[CLS]$ output and $h_t$ is the $t$th hidden state generated from LSTM. $Q_i$, $Q_f$, $Q_c$, $Q_o$, $U_i$, $U_f$, $U_c$ and $U_o$ are weight matrices, and $b_i$, $b_f$, $b_c$, and $b_o$ are bias vectors. 
 
 The attention pooling operation we use is similar to the work of~\citep{Dong:2017}, which is defined as follows:

\begin{center}
\label{eq:atten_alpha_hat}
$\hat \alpha_t=tanh(Q_a \cdot h_t+b_a)$ \\
\label{eq:atten_alpha}
$\alpha_t=\frac{e^{q_a \cdot \hat \alpha_t}}{\sum_{j} e^{q_a \cdot \hat \alpha_j}}$ \\
\label{eq:atten_o}
$o=\sum_{t}{\alpha_t \cdot h_t}$ \\
\end{center}

 $o$ is the segment-scale essay representation corresponding to the scale $k_i$.
 $\alpha_t$ is the attention weight for hidden state $h_t$.
 $Q_a$, $b_a$, $q_a$ are the weight matrix, bias and weight vector respectively.

\subsection{Model Architecture}
\begin{figure*}[ht]
\centering
\includegraphics[width = .95\textwidth]{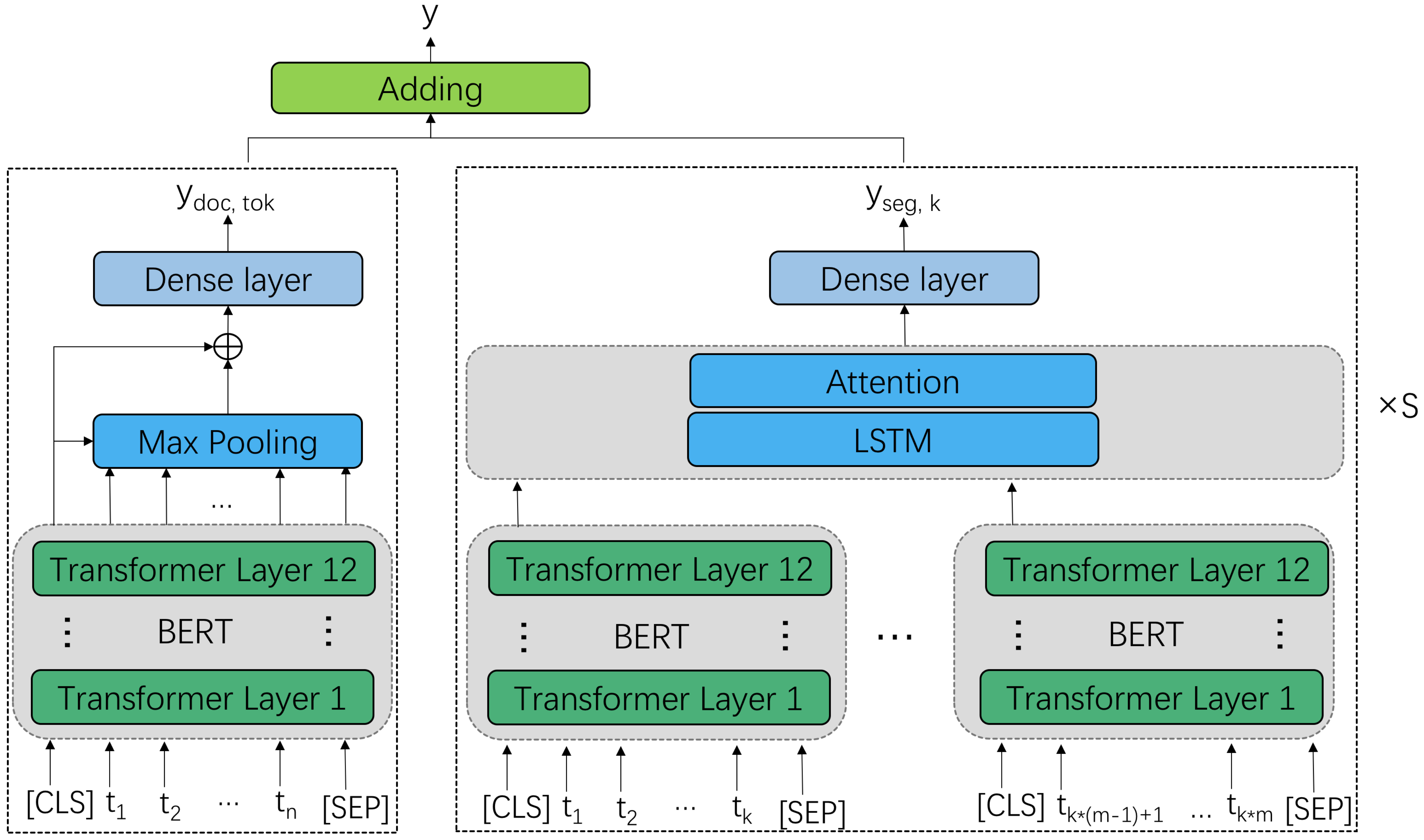}
\caption{
The proposed automated essay scoring architecture based on multi-scale essay representation.
The left part illustrates the document-scale and token-scale essay representation and scoring module,
and the right part illustrates $S$ segment-scale essay representations and scoring modules.}
\label{fig:framework}
\end{figure*}

The model architecture is depicted in Figure~\ref{fig:framework}.

We apply one BERT model to obtain the document-scale and token-scale essay representation.
The concatenation of them is input into a dense regression layer which predicts the score corresponding to the document-scale and token-scale.
For each segment-scale $k$ with number of segments $m$, we apply another BERT model to get $m$ $CLS$ outputs, and apply an $LSTM$ model followed by an attention layer to get the segment-scale representation.
We input the segment-scale representation into another dense regression layer to get the score corresponding to segment-scale $k$.
The final score is obtained by adding the scores of all $S$ segment-scales and the score of the document-scale and token-scale, which is illustrated as below:
 

\begin{center}
\label{eq:pred_score}
$ y= \sum_{k} y_k + y_{doc, tok} $  \\
\label{eq:pred_score_seg}
$ y_{k} = \hat W_{seg} \cdot o_k + b_{seg} $ \\
\label{eq:pred_score_doc_tok}
$ y_{doc, tok} = \hat W_{doc, tok} \cdot H_{doc, tok} + b_{doc, tok}$ \\
\label{eq:pred_score_doc_tok}
$ H_{doc, tok}= w_{doc} \bigoplus W $\\
\end{center}

$y_k $ is the predicted score corresponding to segment-scale $k$.
$y_{doc, tok}$ is the predicted score corresponding to the document-scale and token-scale.
$\hat W_{seg}$ and $b_{seg}$ are weight matrix and bias for segment-scale respectively.
$W_{doc, tok}$ and $b_{doc, tok}$ are weight matrix and bias for document and token-scales, $o_k$ is the segment-scale essay representation with the scale $k$.
$w_{doc}$ is the document-scale essay representation.
$W$ is the token-scale essay representation.
$H_{doc, tok}$ is the concatenation of document-scale and token-scale essay representations.

\subsection{Loss Function}
We use three loss functions to train the model. 

\textbf{MSE} measures the average value of square errors between predicted scores and labels, which is defined as below:

\begin{center}
\label{eq:mse}
$MSE(y, \hat y)=\frac{1}{N} \sum_{i} (y_i-\hat y_i)^2$
\end{center}

where $y_i$ and $\hat y_i$ are the predicted score and the label for the $i$th essay respectively, $N$ is the number of the essays.
 
\textbf{Similarity (SIM)} measures whether two vectors are similar or dissimilar by using cosine function.
A teacher takes into account the overall level distribution of all the students when rating an essay.
Following such intuition, we introduce the SIM loss to the AES task.
In each training step, we take the predicted scores of the essays in the batch as the predicted vector $y$, and the labels as the label vector $\hat y$.
The SIM loss awards the similar vector pairs to make the model think more about the correlation among the batch of essays.
The SIM loss is defined as below:

\begin{center}
\label{eq:sim}
$SIM(y, \hat y)= 1 - cos(y, \hat y)$\\
$y=[y_1, y_2, ..., y_N]$ \\
$\hat y=[\hat y_1, \hat y_2, ..., \hat y_N]$ \\
\end{center}

where $y_i$ and $\hat y_i$ are the predicted score and label for the $i$th essay respectively, $N$ is the number of the essays.

\textbf{Margin Ranking (MR)} measures the ranking orders for each essay pair in the batch.
We intuitively introduce MR loss because the sorting property between essays is a key factor to scoring.
For each batch of essays, we first enumerate all the essay pairs, and then compute the MR loss as follows.
The MR loss attempts to make the model penalize wrong order.
\begin{center}
$ MR(y, \hat y) = \frac{1}{\hat N} {\sum_{i,j} max(0, -r_{i,j}( y_{i} - y_{j}) + b)} $
$ r_{i,j}=\left\{
\begin{array}{lcl}
$1$    &     & { \hat y_i > \hat y_j}\\
$-1$    &     & { \hat y_i < \hat y_j}\\
$-$ sgn(y_i - y_j$)$    &     & { \hat y_i = \hat y_j}\\
\end{array} \right. $
\end{center}
$y_i$ and $\hat y_i$ are the predicted score and label for the $i$th essay respectively.
$\hat N$ is the number of the essay pairs.
$b$ is a hyper parameter, which is set to 0 in our experiment.
For each sample pair $(i, j)$,  when the label $\hat y_i$ is larger than $\hat y_j$, the predicted result $y_i$ should be larger than $y_j$, otherwise, the pair contributes $y_j-y_i$ to the loss. When $\hat y_i$ is equal to $\hat y_j$, the loss is actually $\left| y_i - y_j \right|$.

The combined loss is described as below:

$Loss_{total}(y,\hat y)=\alpha MSE(y, \hat y) + \beta MR(y, \hat y) + \gamma SIM(y, \hat y)$. 

$\alpha$, $\beta$, $\gamma$ are weight parameters which are tuned according to the performance on develop set.

\section{Experiment}

\subsection{Data and Evaluation}

\textbf{ASAP} data set is widely used in the AES task, which contains eight different prompts.
A detailed description can be seen in Table~\ref{font-t1}. For each prompt, the WordPiece length indicates the smallest number which is bigger than the length of 90\% of the essays in terms of WordPiece number.
We evaluate the scoring performance using QWK on ASAP data set, which is the official metric in the ASAP competition. Following previous work, we adopt 5-fold cross validation with 60/20/20 split for train, develop and test sets.
\begin{table}[b!]
\begin{center}
\scalebox{0.7}{
\begin{tabular}{c|cccc}
\hline \bf Prompt& \bf Essays& \bf Avg length&  \bf  Score Range &  \bf  WordPiece length\\ \hline
1 & 1783 & 350 & 2-12 &649\\
2 & 1800 & 350 & 1-6 &704\\
3 & 1726 & 150 & 0-3 &219\\
4 & 1772 & 150 & 0-3 &203\\
5 & 1805 & 150 & 0-4 & 258\\
6 & 1800 & 150 & 0-4 & 289\\
7 & 1569 & 250 & 0-30 & 371\\
8 & 723 & 650 & 0-60 & 1077\\
\hline
\end{tabular}}
\end{center}
\caption{\label{font-t1} Statistics of ASAP data set. }
\end{table}

\textbf{CRP} data set provides 2834 excerpts from several time periods and reading ease scores which range from -3.68 to 1.72. The average length of the excerpts is 175 and the WordPiece length is 252. We also use 5-fold cross validation with 60/20/20 split for train, develop and test sets on CRP data set. As the RMSE metric is used in the CRP competition, we also use it to evaluate our system in ease score prediction task.

\subsection{Baseline}
The baseline models for comparison are described as follows.

\textbf{EASE}~\footnote{\url{http://github.com/edx/ease}} is the best open-source system that participated in the ASAP competition and ranked the third place among 154 participants.
EASE uses regression techniques with handcrafted features.
Results of EASE with the settings of Support Vector Regression (SVR) and Bayesian Linear Ridge Regression (BLRR) are reported in~\citep{Phandi:2015}.

\textbf{CNN+RNN}~Various deep neural networks based on CNN and RNN for AES are studied by~\citep{Taghipour:2016}.
They combine CNN ensembles and LSTM ensembles over 10 runs and get the best result in their experiment.

\textbf{Hierarchical LSTM-CNN-Attention}~\citep{Dong:2017} builds a hierarchical sentence-document model, which uses CNN to encode sentences and LSTM to encode texts.
The attention mechanism is used to automatically determine the relative weights of words and sentences in generating sentence representations and text representations respectively. They obtain the state-of-the-art result among all neural models without pre-training. 

\textbf{SKIPFLOW}~\citep{Tay:2018} proposes to use SKIPFLOW mechanism to model the relationships between snapshots of the hidden representations of an LSTM.
The work of ~\citep{Tay:2018} also obtains the state-of-the-art result among all neural models without pre-training. 

\textbf{Dilated LSTM with Reinforcement Learning}~\citep{Wang:2018} proposes a method using a dilated LSTM network in a reinforcement learning framework.
They attempt to directly optimize the model using the QWK metric which considers the rating schema.

\textbf{HA-LSTM+SST+DAT and BERT+SST+DAT} ~\citep{Cao:2020} propose to use two self-supervised tasks and a domain adversarial training technique to optimize their training, which is the first work to use pre-trained language model to outperform LSTM based methods. They experiment with both hierarchical LSTM model and BERT in their work, which are $HA-LSTM+SST+DAT$ and $BERT+SST+DAT$ respectively.

\textbf{BERT$^2$} ~\citep{Yang:2020} combines regression and ranking to fine-tune BERT model which also outperforms LSTM based methods and even obtains the new state-of-the-art.


\subsection{Settings}
To compare with the baseline models and further study the effectiveness of multi-scale essay representations, losses and transfer learning, we conduct the following experiments. 

\textbf{Multi-scale Models}. These models are optimized with MSE loss, and 
\textbf{BERT-DOC} represents essays with document-scale features based on BERT.
\textbf{BERT-TOK} represents essays with token-scale features based on BERT.
\textbf{BERT-DOC-TOK} represents essays with both document-scale and token-scale features based on BERT.
\textbf{BERT-DOC-TOK-SEG} represents essays with document-scale, token-scale, and multiple segment-scale features based on BERT.
Longformer~\citep{Beltagy2020} is an extension for transformers with an attention mechanism that scales linearly with sequence length, making it easy to process long documents.
We conduct experiments to show that our multi-scale features also works with Longformer and can further improve the performance in long text tasks.
\textbf{Longformer-DOC-TOK-SEG} uses document-scale, token-scale, and multiple segment-scale features to represent essays, but based on Longformer instead of BERT.
\textbf{Longformer-DOC} represents essays with document-scale features based on Longformer. 

\textbf{Models with Transfer Learning}.
To transfer learn from the out-of-domain essays~\footnote{For each prompt, we use all the essays from other prompts in ASAP data set.}, we additionally employ a pre-training stage, which is similar to the work of~\citep{Song:2020}.
In this stage, we scale all the labels of essays from out-of-domain data into range 0-1 and pre-train the model on them with MSE loss.
After the pre-training stage, we continue to fine-tune the model on in-domain essays.
\textbf{Tran-BERT-MS} has the same modules as \textbf{BERT-DOC-TOK-SEG} with pre-training on out-of-domain data.
\textbf{MS} means multiple scale features.

\begin{table*}[h!]
\begin{center}
\scalebox{0.65}{
\begin{tabular}{|c|c|cccccccc|c|}
\hline \bf ID&\bf Models& \bf P1& \bf P2& \bf P3& \bf P4& \bf P5& \bf P6& \bf P7& \bf P8& \bf Average \\ \hline
1&EASE(SVR) \citep{Phandi:2015} & 0.781  &0.621 & 0.630 &0.749 &0.782 &0.771 &0.727 &0.534  &0.699 \\
2&EASE(BLRR) \citep{Phandi:2015} & 0.761 &0.606 & 0.621 &0.742& 0.784& 0.775 &0.730 &0.617 &  0.705\\
\hline
3&CNN(10 runs) + LSTM(10 runs) \citep{Taghipour:2016} & 0.821& 0.688 &0.694& 0.805 & 0.807& 0.819&0.808&0.644 &0.761\\
\hline
4&Hierarchical LSTM-CNN-Attention \citep{Dong:2017}& 0.822 & 0.682 & 0.672&0.814$^*$&0.803&0.811&0.801&0.705&0.764\\
\hline
5&SKIPFLOW LSTM(Bilinear) \citep{Tay:2018} &0.830 &0.678& 0.677& 0.778& 0.795& 0.807& 0.790& 0.670 & 0.753 \\
6&SKIPFLOW LSTM(Tensor) \citep{Tay:2018} & 0.832 &0.684 & 0.695 &0.788& 0.815& 0.810& 0.800 &0.697 & 0.764\\
\hline
7&Dilated LSTM With RL \citep{Wang:2018}& 0.776 & 0.659 & 0.688&0.778&0.805&0.791&0.760&0.545&0.724\\
\hline
8&HA-LSTM+SST+DAT \citep{Cao:2020}& \bf 0.836 & \bf{0.730} & \bf{0.732}&0.822& 0.835 &0.832$^*$&0.821&0.718&0.790\\
9&BERT+SST+DAT \citep{Cao:2020}& 0.824 & 0.699 & \bf 0.726&\bf{0.859}&0.822&0.828&\bf 0.840&0.726& 0.791$^*$\\
\hline
10 & R$^2$BERT \citep{Yang:2020}&0.817& 0.719&0.698&0.845&\bf0.841&\bf \bf 0.847&0.839& 0.744& \bf 0.794$^*$\\
\hline
11& \bf BERT-DOC-TOK-SEG & \bf 0.836 & 0.695 & 0.700 & 0.815 & 0.812 & 0.816 & 0.838 & 0.744 &0.782\\
12 & \bf Tran-BERT-MS-ML-R &0.834&0.716&0.714&0.812&0.813& 0.836&0.839&\bf0.766& 0.791$^*$\\
\hline

\hline
\end{tabular}}
\end{center}
\caption{\label{font-t2} Experiment results of all models in terms of QWK on ASAP. The name of our implemented models are in bold. The bold number is the best performance for each prompt. The best 3 average QWK are annotated with $^*$.}
\end{table*}

\begin{table}[h!]
\begin{center}
\scalebox{0.65}{
\begin{tabular}{|c|c|ccc|c|}
\hline \bf ID&\bf Models& \bf P1& \bf P2& \bf P8& \bf Average \\ \hline
8&HA-LSTM+SST+DAT& 0.836 & 0.730 &0.718&0.761 \\
9&HA-BERT+SST+DAT& 0.824 & 0.699 &0.726&0.750 \\
\hline
10&R$^2$BERT& 0.817 & 0.719 & 0.744 & 0.760\\
\hline
12&\bf Tran-BERT-MS-ML-R&0.834&0.716& 0.766&0.772\\
\hline
\end{tabular}}
\end{center}
\caption{\label{font-t3} Experiment results of our model and the state-of-the-art models on ASAP long essays (WordPiece length are longer than 510). The name of our implemented model is in bold.}
\end{table}

\textbf{Models with Multiple Losses}.
Based on \textbf{Tran-BERT-MS} model, we explore the performance of adding multiple loss functions. \textbf{Tran-BERT-MS-ML} additionally employs MR loss and SIM loss.
\textbf{ML} means multiple losses.
\textbf{Tran-BERT-MS-ML-R} incorporates R-Drop strategy~\citep{Liang:2021} in training based on~\textbf{Tran-BERT-MS-ML} model. 
 


For the proposed model architecture which is depicted in Figure~\ref{fig:framework}, the BERT model in the left part are shared by the document-scale and token-scale essay representations,
and the other BERT model in the right part are shared by all segment-scale essay representations.
We use the "bert-base-uncased" which includes 12 transformer layers and the hidden size is 768.
In the training stage, we freeze all the layers in the BERT models except the last layer, which is more task related than other layers.
The Longformer model used in our work is "longformer-base-4096".
For the MR loss, we set $b$ to 0.
The weights $\alpha$, $\beta$ and $\gamma$ are tuned according to the performance on develop set.
We use Adam optimizer~\citep{Kingma:2015} to fine-tune model parameters in an end-to-end fashion with learning rate of 6e-5, $\beta1$=0.9, $\beta2$=0.999, $L2$ weight decay of 0.005.
The coefficient weight $\alpha$ in R-Drop is 9.
We set the batch size to 32.
We use dropout in the training stage and the drop rate is set to 0.1.
We train all the models for 80 epochs, and select the best model according the performance on the develop set.
We use a greedy search method to find the best combination of segment scales, which is shown in detail in Appendix~\ref{sec:appendix}.
Following~\citep{Cao:2020}, we perform the significance test for our models.

\subsection{Results}

Table~\ref{font-t2} shows the performance of baseline models and our proposed models with joint learning of multi-scale essay representation. Table~\ref{font-t3} shows the results of our model and the state-of-the-art models on essays in prompt 1, 2 and 8, whose WordPiece length are longer than 510.
We summarize some findings from the experiment results.

\begin{itemize}
\item Our model 12 almost obtains the published state-of-the-art for neural approaches.
For the prompts 1,2 and 8, whose WordPiece length are longer than 510, we improve the result from 0.761 to 0.772. As Longformer is good at encoding long text, we also use it to encode essays of prompt 1, 2 and 8 directly but the performance is poor compared to the methods in Table~\ref{font-t3}. 
The results demonstrate the effectiveness of the proposed framework for encoding and scoring essays. We further re-implement BERT$^2$ proposed by ~\citep{Yang:2020}, and our implementation of BERT$^2$ is not as well-performing as the published result. Though ~\citep{Uto:2020} obtain a much better result(QWK 0.801), our method performs much better than their system with only neural features(QWK 0.730), which demonstrates the strong essay encoding ability of our neural approach.

\item Compared to the models 4 and 6, our model 11 uses multi-scale features to encode essays instead of LSTM based models, and we use the same regression loss to optimize the model.
Our model simply changes the representation way and significantly improves the result from 0.764 to 0.782, which demonstrates the strong encoding ability armed by multi-scale representation for long text. Before that, the conventional way of using BERT can not surpass the performance of models 4 and 6. 


\end{itemize}

\subsection{Further analysis}



\textbf{Multi-scale Representation}
We further analyze the effectiveness of employing each scale essay representation to the joint learning process.

\begin{table}[h!]
\begin{center}
\scalebox{0.65}{
\begin{tabular}{|c|c|}
\hline \bf Models& \bf Average QWK \\ \hline
BERT-DOC & 0.760\\
\hline
BERT-TOK & 0.764\\
\hline
BERT-DOC-TOK & 0.768\\
\hline
BERT-DOC-TOK-SEG &0.782\\
\hline
\end{tabular}}
\caption{\label{font-t4} Performance of different feature scale models on ASAP data set.}
\end{center}
\end{table}

\begin{table}[h!]
\begin{center}
\scalebox{0.65}{
\begin{tabular}{|c|c|}
\hline \bf Models& \bf RMSE \\ \hline
BERT-DOC & 0.742\\
\hline
BERT-TOK &  0.760\\
\hline
BERT-DOC-TOK & 0.691\\
\hline
BERT-DOC-TOK-SEG & 0.607 \\
\hline
\end{tabular}}
\caption{\label{font-t5} 
Performance of different feature scale models on CRP data set.
The evaluation metric is RMSE.
Lower numbers are better.}
\end{center}
\end{table}

Table~\ref{font-t4} and Table~\ref{font-t5} show the performance of our models to represent essays on different feature scales, which are trained with MSE loss and without transfer learning.
Table~\ref{font-t4} shows the performance on ASAP data set while Table~\ref{font-t5} shows the performance on CRP data set. The improvement of BERT-DOC-TOK-SEG over BERT-DOC, BERT-TOK, BERT-DOC-TOK are significant ($p$<0.0001) on CRP data set, and are significant ($p$<0.0001) in most cases on ASAP data set.
Results on both table indicate the similar findings.
\begin{itemize}

\item Combining the features from document-scale and token-scale, BERT-DOC-TOK outperforms the models BERT-DOC and BERT-TOK, which only use one scale features.
This demonstrates that our proposed framework can benefit from multi-scale essay representation even with only two scales.

\item By additionally incorporating multiple segment-scale features, BERT-DOC-TOK-SEG performs much better than BERT-DOC-TOK.
This demonstrates the effectiveness and generalization ability of our multi-scale essay representation on multiple tasks.

\begin{table}[h!]
\begin{center}
\scalebox{0.65}{
\begin{tabular}{|c|c|}
\hline \bf Models& \bf Average QWK \\ \hline
Longformer-DOC&0.746\\
\hline
Longformer-DOC-TOK-SEG&0.771\\
\hline
\end{tabular}}
\caption{\label{font-t6} Performance of multi-scale Longformer models on ASAP data set.}
\end{center}
\end{table}

\end{itemize}

\textbf{Reasons for Effectiveness of Multi-scale Representation}
Though the experiment shows the effectiveness of multi-scale representation, we further explore the reason.
We could doubt that the effectiveness comes from supporting long sequences, not the multi-scale itself.
As Longformer is good at dealing with long texts, we compare the results between Longformer-DOC and Longformer-DOC-TOK-SEG. The results of the significance test show that the improvement of Longformer-DOC-TOK-SEG over Longformer-DOC are significant ($p$<0.0001) in most cases. Performance of the two models are shown in Table~\ref{font-t6}, and we get the following findings. 

\begin{itemize}
\item Though Longformer-DOC supports long sequences encoding, it performs poor, which indicates us that supporting long sequence ability is not enough for a good essay scoring system.

\item Longformer-DOC-TOK-SEG outperforms Longformer-DOC significantly, which indicates the effectiveness of our model comes from encoding essays by multi-scale features, not only comes from the ability to deal with long texts.
\end{itemize}

These results are consistent with our intuition that our approach takes into account different level features of essays and predict the scores more accurately. 
We consider it caused by that multi-scale features are not effectively constructed in the representation layer of pre-trained model due to the lack of data for fine-tuning in the AES task.
Therefore, we need to explicitly model the multi-scale information of the essay data and combine it with the powerful linguistic knowledge of pre-trained model.

\begin{table}[h!]
\begin{center}
\scalebox{0.65}{
\begin{tabular}{|c|c|}
\hline \bf Models& \bf Average \\ \hline
BERT-DOC-TOK-SEG& 0.782 \\
Tran-BERT-MS& 0.788 \\
Tran-BERT-MS-ML& 0.790 \\
Tran-BERT-MS-ML-R& 0.791 \\
\hline
\end{tabular}}
\end{center}
\caption{\label{font-t7} Experiment results for transfer learning with multiple loss functions and R-Drop .}
\end{table}

\textbf{Transfer Learning with Multiple Losses and R-Drop}
We further explore the effectiveness of pre-training with adding multiple loss functions and employing R-Drop. 
As is shown in table~\ref{font-t7}, by incorporating the pre-training stage which learns the knowledge from out-of-domain data, Tran-BERT-MS model improves the result from 0.782 to 0.788 compared to BERT-DOC-TOK-SEG model.
The model Tran-BERT-MS-ML which jointly learns with multiple loss functions further improves the performance from 0.788 to 0.790.
We consider it due to the reason that MR brings ranking information and SIM takes into account the overall score distribution information.
Diverse losses bring different but positive influence on the optimization direction and act as an ensembler.
By employing R-Drop, Tran-BERT-MS-ML-R improves the QWK slightly, which comes from the fact that R-Drop plays a regularization role.

\section{Conclusion and Future Work}
In this paper, we propose a novel multi-scale essay representation approach based on pre-trained language model, and employ multiple losses and transfer learning for AES task.
We almost obtain the state-of-the-art result among deep learning models.
In addition, we show multi-scale representation has a significant advantage when dealing with long texts.

One of the future directions could be exploring soft multi-scale representation.
Introducing linguistic knowledge to segment at a more reasonable scale may bring further improvement.

\bibliography{anthology,custom}
\bibliographystyle{acl_natbib}

\appendix

\section{Appendix}
\label{sec:appendix}

All the segment-scales we explore range from 10 to 190.
The interval between two neighbor scales is 20. As the combination number of all segment-scales is exponential, we use a greedy search method to find the best combination.
\begin{enumerate}
\item Initialize the segment-scale value set $R$ as the document-scale and  token-scale.
\item Experiment the combination of each segment-scale with the token-scale and document-scale essay representation, and compute the average QWK on develop set for all segment-scales, which is denoted as $QWK_{ave}$.
The scale with higher QWK compared to $QWK_{ave}$ is added to the candidate scale list $L$ and the scales in $L$ are sorted according to their QWK values from large to small.
\item For each $i$ from 1 to $|L|$, we perform experiments on the combination of the first $i$ segment-scales in $L$ with the token-scale and document-scale.
The combination segment-scales with the best performance on develop set are added to the segment-scale value set $R$
\end{enumerate}

\end{document}